\documentclass[final,5p,times,twocolumn]{elsarticle}%review

\usepackage{lineno,hyperref}
\modulolinenumbers[5]
\journal{Pattern Recognition Letters}

\usepackage[utf8]{inputenc} % allow utf-8 input
\usepackage[T1]{fontenc}    % use 8-bit T1 fonts
\usepackage[]{hyperref}       % hyperlinks
\usepackage{url}            % simple URL typesetting
\usepackage{booktabs}       % professional-quality tables
\usepackage{amsfonts}       % blackboard math symbols
\usepackage{nicefrac}       % compact symbols for 1/2, etc.
\usepackage{microtype}      % microtypography
\usepackage{xcolor}         % colors

\usepackage{graphicx}
\usepackage{subfigure}
\usepackage{multirow}
\usepackage{romannum}
\usepackage[export]{adjustbox}
\usepackage{enumerate}
\usepackage{amsmath,amssymb,amsthm,mathtools,nccmath}
\usepackage{caption}
\usepackage{dsfont}
\usepackage{textcomp}
\usepackage[multiple]{footmisc}
\usepackage{algorithm,algorithmic}
\usepackage{wrapfig}
\usepackage{enumitem}
\usepackage{grffile}

\theoremstyle{definition}
\newtheorem{definition}{Definition}
\newtheorem*{definition*}{Definition}

\newtheorem*{example*}{Example}
\theoremstyle{theorem}

\newtheorem{proposition}[definition]{Proposition}

\newtheorem*{theorem*}{Theorem}
\newtheorem*{lemma*}{Lemma}
\newtheorem*{proposition*}{Proposition}
\newtheorem*{corollary*}{Corollary}
\newtheorem*{remark*}{Remark}
\newtheorem*{claim*}{Claim}
\newtheorem*{problem*}{Problem}
\newtheorem*{observation*}{Observation}

\allowdisplaybreaks

\begin{document}
\pagenumbering{arabic}

\begin{frontmatter}

\title{Generalized Gumbel-Softmax Gradient Estimator for Generic Discrete Random Variables}
%\tnotetext[mytitlenote]{Fully documented templates are available in the elsarticle package on \href{http://www.ctan.org/tex-archive/macros/latex/contrib/elsarticle}{CTAN}.}

%% Group authors per affiliation:
\author[add1]{Weonyoung Joo\corref{cor}}
\ead{weonyoungjoo@ewha.ac.kr}

\author[add2]{Dongjun Kim}
\ead{dongjoun57@kaist.ac.kr}

\author[add2]{Seungjae Shin}
\ead{tmdwo0910@kaist.ac.kr}

\author[add2]{Il-Chul Moon}
\ead{icmoon@kaist.ac.kr}

\address[add1]{Department of Statistics, EWHA Womans University, Seoul, Republic of Korea}
\address[add2]{Department of Industrial and Systems Engineering, Korea Advanced Institute of Science and Technology, Daejeon, Republic of Korea}
\cortext[cor]{Corresponding author}

\begin{abstract}
Estimating the gradients of stochastic nodes in stochastic computational graphs is one of the crucial research questions in the deep generative modeling community, which enables the gradient descent optimization on neural network parameters. 
%Estimating the gradients of stochastic nodes, which enables the gradient descent optimization on neural network parameters, is one of the crucial research questions in the deep generative modeling community. 
Stochastic gradient estimators of discrete random variables are widely explored, for example, Gumbel-Softmax reparameterization trick for Bernoulli and categorical distributions. 
Meanwhile, other discrete distribution cases such as the Poisson, geometric, binomial, multinomial, negative binomial, etc. have not been explored. 
This paper proposes a generalized version of the Gumbel-Softmax estimator, which is able to reparameterize generic discrete distributions, not restricted to the Bernoulli and the categorical. 
The proposed estimator utilizes the truncation of discrete random variables, the Gumbel-Softmax trick, and a special form of linear transformation. 
%, and it enables learning on a large-scale stochastic computational graph. 
Our experiments consist of (1) synthetic examples and applications on VAE, which show the efficacy of our methods; and (2) topic models, which demonstrate the value of the proposed estimation in practice.
\end{abstract}

\begin{keyword}
Reparameterization Trick \sep Discrete Random Variable \sep Gumbel-Softmax Trick \sep Variational Autoencoder \sep Deep Generative Model
\end{keyword}

\end{frontmatter}

\section{Introduction}\label{introduction}

Stochastic computational graphs, including variational autoencoders (VAEs) \citep{Kingma14b}, are widely used for probabilistic modeling, representation learning, and generating data in the deep generative model (DGM) society. %\citep{Jiang19}. % Song22
Optimizing the network parameters through back-propagating gradients requires an estimation of the gradient values. 
However, the stochasticity requires the computation of expectation, which differentiates this problem from the deterministic gradient of ordinary neural networks. 
Regarding such a perspective, there are two common ways of estimating the gradients from stochastic nodes: score function (SF) methods and reparameterization methods.
The SF-based estimators tend to result in unbiased gradients with high variances, hence, the SF-based estimators aim to reduce the variances of gradients for stable and fast optimizations.
Meanwhile, the reparameterization estimators result in biased gradients with low variances \citep{Xu19}, but they require the \textit{differentiable non-centered parameterization} \citep{Kingma14a} of random variables.

For continuous random variables such as the Gaussian, the reparameterization estimators are widely utilized due to the nature of the differentiability of the typical continuous distributions \citep{Kingma14b,Nalisnick17,Joo19}. 
Nowadays, it is feasible to estimate gradients for continuous cases with automatic differentiation \citep{Figurnov18,Jankowiak18} in TensorFlow \citep{Abadi16} or PyTorch \citep{Paszke19}.
Meanwhile, for the discrete random variables which follow Bernoulli or categorical distributions, the SF-based methods are widely explored since the reparameterization method can not directly work due to the non-differentiability.
Alternatively, the Gumbel-Softmax trick \citep{Jang17,Maddison17} overcomes this difficulty through the reparameterization with continuous relaxation of one-hot selection values.
Also, a line of works \citep{Tucker17,Grathwohl17} utilizes both the SF-based method and the reparameterization method.

%For random variables which follow continuous distributions, the reparameterization estimators are widely used due to the nature of differentiability of the popular continuous distributions.
%Gaussian VAE \citep{Kingma14b} utilizes the exact reparameterization form, while other VAEs with explicit priors suggest the reparameterization tricks with approximations \citep{Nalisnick17,Joo19}.
%Also, it is feasible to estimate gradients for the continuous cases with automatic differentiation \citep{Figurnov18,Jankowiak18} in TensorFlow \citep{Abadi16} or PyTorch \citep{Paszke19}.

%On the other hand, for the discrete random variables following Bernoulli or categorical distributions, the SF-based methods are well-explored since the reparameterization method does not directly work due to the non-differentiability.
%Meanwhile, Gumbel-Softmax trick \citep{Jang17,Maddison17} overcome the difficulty from the non-differentiability of discrete distributions, through reparameterization by continuous relaxation of one-hot selection value.
%Also, there is a line of works which utilize both the SF-based method and the reparameterization method such as \citet{Tucker17,Grathwohl17}.

While the Bernoulli and the categorical cases have been studied deeply, the gradient estimators for other discrete distributions have hardly been explored.
There are several gradient estimators which are specialized in discrete distributions on combinatorial spaces, such as $k$-hot vector, permutation, and spanning trees on graphs \citep{Gadetsky20,Paulus20}.
However, other renowned discrete distribution cases, such as Poisson, binomial, multinomial, geometric, negative binomial distributions, etc., have not been explored, which we mainly focus on in this paper.
Prior works on probabilistic graphical models, such as \citet{Ranganath15,Ranganath16}, adopted Poisson latent variables for latent counting.
Another line of work \citep{Wu20} utilized the Gaussian approximation on the Poisson to count the number of words in deep generative modeling, which can be a poor approximation when the rate parameter is small.
Regarding those perspectives, the stochastic gradient estimator for the discrete distribution needs to be studied further to extend the choice of prior assumptions.
%Then, we can migrate the ideas on the probabilistic graphical models to the DGMs with the utilization of various distributions.

This paper proposes a generalized version of the Gumbel-Softmax trick, which can reparameterize the broader ranges of discrete distributions, not limited to the Bernoulli and the categorical.
The proposed Generalized Gumbel-Softmax (\textsc{GenGS}) is probabilistically grounded, and it utilizes (1) a truncation to finitize the infinite supports of the discrete distributions; and (2) a transformation that enables the generalization of the Gumbel-Softmax trick.
%\textsc{GenGS} is probabilistically grounded in that the reparameterization estimator with continuous relaxation approximates the broader class of discrete distributions.
%, hence, results in better gradient estimation.
Moreover, through the truncation, \textsc{GenGS} provides a posterior inference procedure that can be either explicit or implicit as a practical implementation. 
Our experiments show the efficacy with synthetic examples and VAEs, as well as the usability in topic model applications.

\begin{figure*}
    \vspace{-.75em}
    \centering
    \subfigure[Reparameterization trick.\label{fig_repara_trick}]{\includegraphics[width=.22\linewidth]{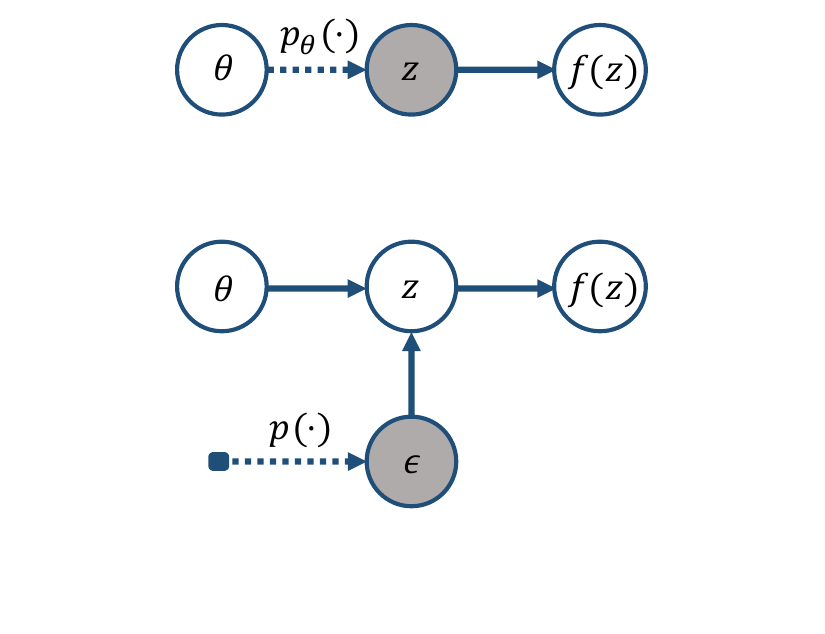}}
    \hspace{1.em}
    \centering
    \subfigure[Visualization of \textsc{GenGS} reparameterization trick.\label{fig_graph_concept}]{\includegraphics[width=.5\linewidth]{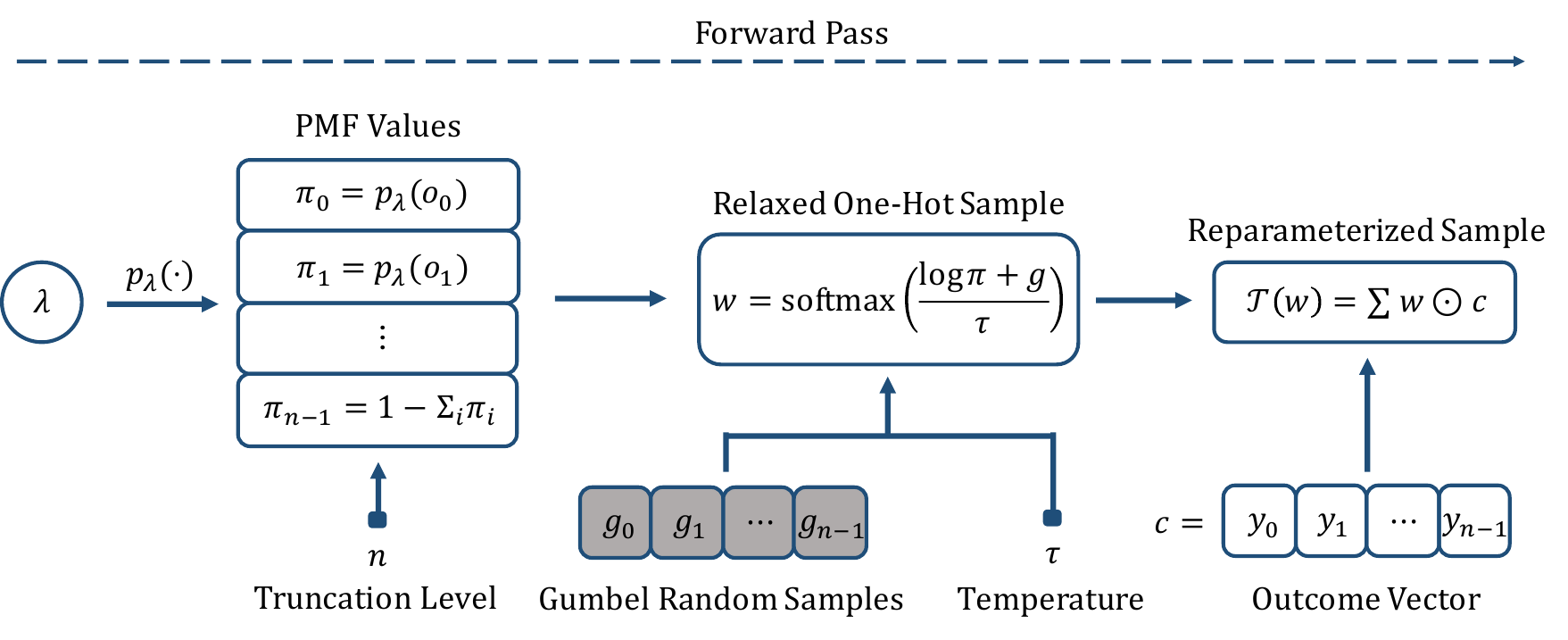}}
    \vspace{-.8em}
    \caption{Visualization of reparameterization trick. The shaded nodes indicate random nodes, and the dotted lines denote sampling processes. The auxiliary random variables enable the back-propagation flow of the gradients.}
    \vspace{-.6em}
\end{figure*}

\section{Preliminary}

\subsection{Back-propagation through Stochastic Nodes}\label{preliminary_bpp_rn}
In the stochastic computational graph, assume that there is an intermediate stochastic node or a latent variable $z \sim p(z|\theta)$, where the distribution depends on the parent node $\theta$ as a distribution parameter.
The goal is optimizing the objective function $\mathcal{L (\theta,\eta)} = \mathds{E}_{z \sim p(z|\theta)} [f_{\eta} (z)]$, where $f_{\eta}$ is a differentiable function with respect to $\eta$, i.e., the neural networks.
To optimize the objective function with respect to the parameter $\theta$ through the gradient methods, we need to compute $\nabla_{\theta} \mathcal{L(\theta,\eta)} = \nabla_{\theta} \mathds{E}_{z \sim p(z|\theta)} [f_{\eta}(z)]$, which is intractable with its original form.

\textit{SF-based estimator} utilizes a score function $\nabla_{\theta} \log p(z|\theta)$ by utilizing log-derivative trick to compute the gradient $\nabla_{\theta} \mathcal{L(\theta,\eta)}$ \citep{Williams92}.
Then, the gradient $\nabla_{\theta} \mathcal{L(\theta,\eta)}$ can be derived as the below:
\begin{align}
    \nabla_{\theta} \mathcal{L} (\theta,\eta) 
    = \mathds{E}_{z \sim p(z|\theta)} [f_{\eta} (z) \nabla_{\theta} \log p(z|\theta)] ~.
\end{align}
The SF methods compute exact gradients due to the unbiasedness with the Monte Carlo property.
However, those methods suffer from the high variance of gradients, which results in the slow and unstable convergence of the objective function.
To reduce the variance of gradients, the control variate methods \citep{Mnih14,Gu16} or the RaoBlackwellization \citep{Liu19,Kool20} are widely used.
%However, these methods still suffer from the high variance of the gradient.
%Recent works focus on the to reduce the variance.

\textit{Reparameterization trick}, illustrated in Figure \ref{fig_repara_trick}, introduces an auxiliary variable $\epsilon \sim p(\epsilon)$, which takes over all randomness of the latent variable $z$, to compute $\nabla_{\theta} \mathcal{L(\theta,\eta)}$.
Then, the sampled value $z$ can be re-written as $z = g(\theta,\epsilon)$, with a \textit{deterministic} and \textit{differentiable} function $g$ in terms of $\theta$.
Here, the gradient $\nabla_{\theta} \mathcal{L(\theta,\eta)}$ is derived as the below:
\begin{align}
    \nabla_{\theta} \mathcal{L} (\theta,\eta) 
     = \mathds{E}_{\epsilon \sim p(\epsilon)} [\nabla_{g} f_{\eta} (g(\theta,\epsilon)) \nabla_{\theta} g(\theta,\epsilon)] ~. \label{test}
\end{align}
Equation \eqref{test} is computable, however, the differentiability requires the continuity of the random variable $z$, so the distribution of $z$ is limited to the continuous distributions.
%where Equation (\ref{test}) is now computable.
%To achieve this form, the reparameterization trick requires the continuity of the random variable $z$, so the distribution of $z$ is limited to the continuous distributions.

\subsection{Gumbel Tricks}\label{preliminary_gumbel}
To utilize the \textit{differentiable} reparameterization trick on discrete random variables, continuous relaxation can be applied.
A \textit{Gumbel-Softmax} (GS) trick \citep{Jang17,Maddison17} is an approximation of a \textit{Gumbel-Max} (GM) trick, which are alternatives of a one-hot categorical sampling.
The categorical random variable $Z \sim \text{Cat} (\pi)$, where $\pi$ in the $(n-1)$-simplex $\Delta^{n-1}$, can be reparameterized by the GM trick: (1) draw $u_{j}\sim \text{Uniform} (0,1)$ to generate a Gumbel sample $g_j = - \log(-\log{u_j})$ for $j = 1,\cdots,n$; and (2) compute $k = \texttt{argmax}_{j=1}^{n} {[g_j + \log{\pi_{j}}]}$.
This procedure generates a one-hot sample $z$ such that $z_{k}=1$ with $P(Z_k = 1) = \pi_k$, and zeros in other entries.
Instead of the \texttt{argmax} in the GM, the GS utilizes the \texttt{softmax} with a temperature $\tau > 0$, i.e., $z \approx \texttt{softmax} \big( \frac{g + \log{\pi}}{\tau} \big)$.
This substitution relaxes the discreteness of the categorical random variable to the one-hot-like form in the continuous domain. %, by approximating the GM with $\tau \rightarrow 0$.
In other words, if we denote $\text{GM}(\pi)$ and $\text{GS}(\pi,\tau)$ as the distribution generated by the GM trick and the GS trick, respectively, then $\text{GS}(\pi,\tau) \rightarrow \text{GM}(\pi)$ as $\tau \rightarrow 0$.
%The GS has been widely used to reparameterize categorical random variables, such as \texttt{RelaxedOneHotCategorical} in TensorFlow or PyTorch.

\section{Methodology}\label{method}
%This section introduces \textsc{GenGS}, the general version of the GS reparameterization trick.
%The key ideas of \textsc{GenGS} are (1) truncating the infinite supports of discrete random variables to approximate the random variable with the finite number of possible outcomes; (2) alternative one-hot sampling with the Gumbel tricks; and (3) relaxation of the categorical selection into the continuous form, via the GS trick.%; and (4) a reversion of the selected category in the relaxed one-hot form to the sample value.

\subsection{Problem Setting}\label{assumption}
We begin an explanation on \textsc{GenGS} with discrete distribution having finite support.
%First, we introduce some mathematical assumptions for further explanation of our methodology.
Assume that a random variable $Z$ follows a discrete distribution $\text{Q}(\lambda)$ with explicit probability mass function (PMF) and a finite support $C = \{ c_1, \cdots, c_n\}$.
% and the corresponding probability mass function (PMF) values $\pi_k = P(Z = c_k)$ for $1 \leq k \leq n < \infty$.
Next, define an outcome vector $c = (c_1, \cdots, c_n)$ and the corresponding PMF value vector $\pi = (\pi_1, \cdots, \pi_n)$ in the same index order.
Here, each sample $c_k$ can be either scalar or vector (or, even matrix) by the choice of distribution $\text{Q}$.
%, and we assume that we have an explicit PMF, hence, we can directly compute PMF values.
Finally, for $w = (w_1, \cdots, w_n) \in \Delta^{n-1}$, introduce a transformation $\mathcal{T}$ as follows:
\begin{align}\label{transformation}
\textstyle \mathcal{T} (w) = \sum_{k=1}^{n} w_k c_k := \sum w \odot c ~.
\end{align}

\subsection{Sampling through Generalized Gumbel Tricks}\label{stohs} % HERE
We claim that the sampling process of $Z$ can be derived by the one-hot categorical selection process, i.e., the GM process.
Suppose that we draw a sample $Z = c_k \in C$, then the sampled value $c_k$ can be alternatively selected as follows.
Since we fix the index order of the outcome vector $c$ and its corresponding PMF vector $\pi$, the one-hot indicating vector of the sample $c_k$ can be regarded as $w = \texttt{one\_hot}(c_k|c)$, which can be alternatively sampled from $\text{GM}(\pi)$.
Then, the remaining step is a conversion of $w = \texttt{one\_hot}(c_k|c)$ to $c_k$, and this step can be done by the linear transformation $\mathcal{T}$ in Equation \eqref{transformation}. 
Hence, we can reparameterize the original sample value $c_k$ as $\mathcal{T} (\text{GM}(\pi))$, as stated in Proposition \ref{thm_td_hgm}.
\begin{proposition}\label{thm_td_hgm}
    For the transformation $\mathcal{T}$, if a discrete random variable $Z \sim \text{Q} (\lambda)$ has a finite support with known PMF values, $Z$ can be reparameterized by $\mathcal{T} (\text{GM}(\pi))$ where $\pi_{k} = P(Z = c_k)$.
\end{proposition}
\begin{proof}
    %Suppose we have a sample $c_m$ from $Z$, and note that we have known PMF values $\pi = (\pi_1, \cdots, \pi_{n})$ of $Z$. 
    With the transformation $\mathcal{T}$ and the constant outcome vector $c = (c_1,\cdots,c_{n})$, the following holds:
    \begin{align*}
    c_m = \textstyle \sum_{k=1}^{n} c_k e_k = \mathcal{T} (e_m) ~.
    \end{align*}
    Since the transformation $\mathcal{T}$ is a deterministic function, by introducing the auxiliary Gumbel random variable, we can remove the randomness of $Z$ from $\lambda$. % with the uniform random variable composing the Gumbel random variable.
    Hence, the discrete random variable $Z$ with finite support can be reparameterized by the GM trick and the transformation $\mathcal{T}$.
\end{proof}

\subsection{Relaxation of One-Hot Sample}\label{roohs}
Replacing the \texttt{argmax} in the GM trick to the \texttt{softmax} enables differentiation, which is a key point of the GS reparameterization.
%Note that the GM process can be continuously relaxed through the GS process.
%The \texttt{softmax} function relaxes the strict one-hot selection vector from the \texttt{argmax} function.
%Also, this replacement enables differentiation which is a key point of the GS reparameterization.
In Proposition \ref{thm_tgs_tgm}, we claim that the same argument still holds under the transformation $\mathcal{T}$: if $W \sim \textsc{GS} (\pi,\tau)$ where $\pi$ is a PMF vector of discrete random variable $Z \sim \text{Q} (\lambda)$ with finite support, then $Z$ can be continuously relaxed and reparameterized as $\mathcal{T} (\textsc{GS} (\pi,\tau))$, which we define as $\textsc{GenGS}(\pi,\tau)$.
%for pre-defined discrete random variable $Z \sim \text{Q} (\lambda)$ with the corresponding PMF vector $\pi$ and the linear transformation $\mathcal{T}$ in Equation (\ref{transformation}), then $\mathcal{T} (W)$ is a continuously relaxed reparameterized version of $Z \sim \text{Q}(\lambda)$.
%Proposition \ref{thm_tgs_tgm} states that the relaxation of \textsc{GenGS} remains the reparameterization under the transformation $\mathcal{T}$.
\begin{proposition}\label{thm_tgs_tgm}
    For the transformation $\mathcal{T}$ and a categorical parameter $\pi \in \Delta^{n-1}$, the convergence property of \textsc{GS} to \textsc{GM} still holds under the linear transformation $\mathcal{T}$, i.e., $\text{GS} (\pi,\tau) \rightarrow \text{GM} (\pi)$ as $\tau \rightarrow 0$ implies $\textsc{GenGS} (\pi,\tau) \rightarrow \mathcal{T} (\text{GM} (\pi))$ as $\tau \rightarrow 0$.
\end{proposition}
\begin{proof}
    %Suppose we have a given categorical parameter $\pi \in \Delta^{n-1}$.
    Define $f_M$ and $f_S^{\tau}$ be GM reparameterization and GS reparameterization with a temperature $\tau > 0$, respectively, where the both take the categorical parameter $\pi \in \Delta^{n-1}$ and a Gumbel sample as inputs.
    %Note that $f_M$ returns a one-hot value, and $f_S^{\tau}$ returns a one-hot-like \texttt{softmax} value.
    For the Gumbel sample $g$, if we assume $g_m + \log \pi_m > g_j + \log \pi_j$ for all $j \neq m$, then the following holds: 
    \begin{align*}
    \mathcal{T} (f_M (\pi, g)) = \textstyle \sum_{k=1}^{n} [e_m]_k \cdot c_k  = \textstyle \sum_{k=1}^{n} e_m \odot c = c_m
    \end{align*}
    where $e_j$ is a $n$-dimensional one-hot vector, which has $1$ in the $j^{\text{th}}$ entry and $0$ in all other entries.
    %i.e., the selected sample as a category is $c_m$ out of possible outcome set $\{ c_1, \cdots, c_{n} \}$.
    %Therefore, for the Gumbel-Max trick, it is clear that $\mathcal{T} (f_M (\pi, g)) = \sum_{k=1}^{n} [e_m]_k \cdot c_k  = \sum_{k=1}^{n} e_m \odot c = c_m$ for $c = (c_1,\cdots,c_{n})$ where $e_j = (0,\cdots,0,1,0,\cdots,0)$ is a $n$-dimensional one-hot vector, which has $1$ in the $j^{\text{th}}$ entry and $0$ in all other entries.
    As $\tau \rightarrow 0$, the statement $\text{GS} (\pi,\tau) \rightarrow \text{GM} (\pi)$ implies $f_S^{\tau} (\pi, g) \rightarrow f_M (\pi, g)$, i.e., 
    \begin{align*}
    \big[ f_S^{\tau} (\pi, g) \big]_j & = \frac{\exp \big( \frac{g_j + \log \pi_j}{\tau} \big) }{\sum_{k=1}^{n} \exp \big( \frac{g_k + \log \pi_k}{\tau} \big) } 
    \rightarrow 
    \begin{cases}
    1 & ~ \text{if} ~ j = m \\
    0 & ~ \text{if} ~ j \neq m
    \end{cases}
    \quad \text{as} ~ \tau \rightarrow 0 ~.
    \end{align*}
    Hence, $f_S^{\tau} (\pi, g) = \tilde{e}_m$ holds for some relaxed one-hot vector of $e_m$ by introducing the softmax relaxation.
    As a consequence, 
    \begin{align*}
    \big[ f_S^{\tau} (\pi, g)\big]_j \times c_j
    & = \Big[ \sum \frac{\exp \big( \frac{g_i + \log \pi_i}{\tau} \big) }{\sum_{k=1}^{n} \exp \big( \frac{g_k + \log \pi_k}{\tau} \big) }\Big]_{j} \times c_j  \\
    & \rightarrow 
    \begin{cases}
    c_m  & ~\text{if} ~ j = m \\
    0 & ~\text{if} ~ j \neq m
    \end{cases}
    \quad \text{as} ~ \tau \rightarrow 0 ~.
    \end{align*}
    Hence, by taking the summation, the following holds:
    \begin{align*}
    \mathcal{T} (f_S^{\tau} (\pi, g)) = \textstyle \sum_{k=1}^{n} [\tilde{e}_m]_k \cdot c_k \rightarrow c_m = \mathcal{T} (f_M (\pi, g)) ~.
    \end{align*}
\end{proof}

\begin{figure*}[t]
    \vspace{-.6em}
    \centering
    \subfigure[Effects of truncation level and temperature\label{fig_grid_of_poisson}]{\includegraphics[width=.49\linewidth]{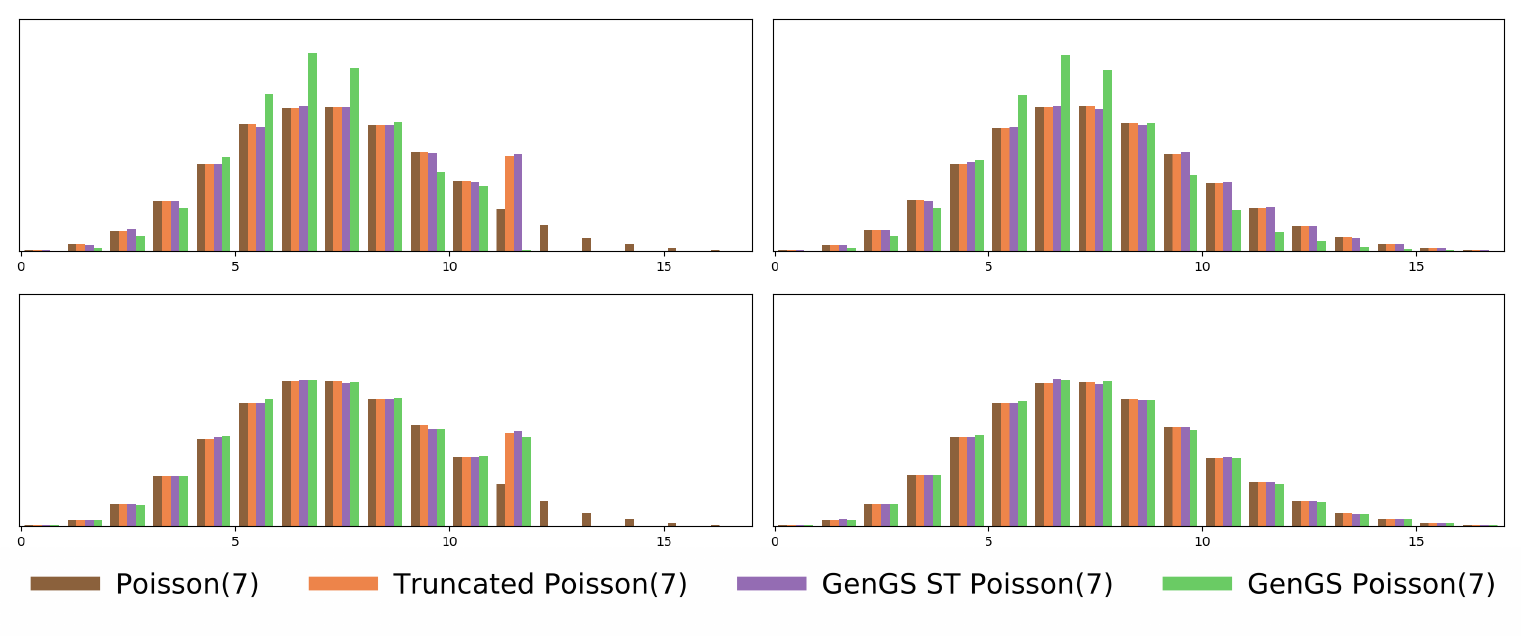}}
    \hspace{3.em}
    \centering
    \subfigure[Concept of \textsc{GenGS} approximation\label{fig_concept}]{\includegraphics[width=.32\linewidth]{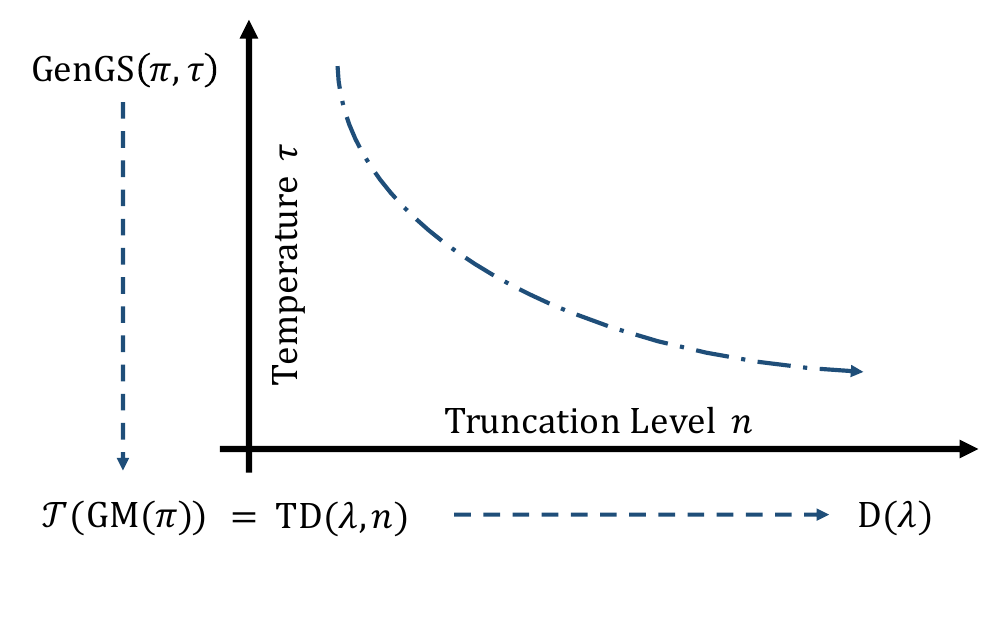}}
    \vspace{-.75em}
    \caption{(a) Approximation of \textsc{GenGS} in terms of choices of the truncation level $n$ and the temperature $\tau$ in $\text{Poisson} (7)$. As sub-figures go from left to right, the truncation level grows. Hence, the popped-out sticks, implying remaining probability of the right side, disappears if the truncation level is large enough. As sub-figures go from top to bottom, the temperature decreases, and the PMF of truncated distributions and the original distributions becomes similar. (b) On the $y$-axis, as temperature $\tau \downarrow 0$, $\textsc{GenGS} (\pi,\tau) \rightarrow \text{TD} (\lambda,n)$, where $\pi$ is a computed PMF value of $\text{TD} (\lambda, n)$, according to Proposition \ref{thm_tgs_tgm}. $\text{TD} (\lambda,n)$ can be reparameterized by the GM trick with a linear transformation $\mathcal{T}$ as in Proposition \ref{thm_td_hgm}. On the $x$-axis, $\text{TD} (\lambda,n) \rightarrow \text{D} (\lambda)$ as truncation level $n \uparrow \infty$, according to Proposition \ref{thm_d_td}.}
    \label{fig_approximation}
    \vspace{-.6em}
\end{figure*}

%\begin{algorithm}[t]
%    \centering
%    \caption{Computing PMF values of truncated distribution $\text{TD} (\lambda,R)$.}
%    \label{algorithm_pmf}
%    \begin{algorithmic}[1]
%    \STATE \textbf{Input:} Distribution $\text{D}(\lambda)$ with parameter $\lambda$ and PMF $p_{\lambda} (\cdot)$, finite support $C = \{ c_1, \cdots, c_n\}$ in truncation range $R$. 
%    \STATE \textbf{for} {$k=1$ {\bfseries to} $n-1$} \textbf{do}
%    \STATE \hspace{.8em} $\pi_k = p_{\lambda} (c_k)$
%    %\ENDFOR 
%    \STATE $\pi_{n} = 1 - \sum_{k=1}^{n-1} \pi_k$
%    \STATE \textbf{return} $\pi=(\pi_1,\cdots,\pi_n)$
%    \end{algorithmic}
%\end{algorithm}

\subsection{Finitizing Support by Truncation}\label{fsbt}
In Section \ref{assumption}, we assumed that the support of the distribution to be finite.
%, hence, we can directly apply the one-hot selecting Gumbel tricks from the finite number of categories.
Without the finite assumption, the Gumbel tricks cannot be applied since it requires a limited number of categories.
Hence, for the discrete distributions with infinite support, we truncate tails of the distributions to approximate the original distribution. 
In that manner, we can extend \textsc{GenGS} to the discrete distribution with infinite support.
Definition \ref{def_truncatable_dist} utilizes truncation range to truncate the non-negative discrete random variable.
%Definition \ref{def_truncatable_dist} and \ref{def_truncatable_dist_both} utilize truncation range to truncate the discrete random variable, where Definition \ref{def_truncatable_dist} is specialized for non-negative distributions and Definition \ref{def_truncatable_dist_both} can be applied to general discrete distributions.
\begin{definition}\label{def_truncatable_dist}
    For a non-negative discrete random variable $X \sim \text{D}(\lambda)$, define $Z_n = X$ if $X \leq n-1$, and $Z_n = n-1$ if $X > n$.
    The random variable $Z_{n}$ is said to follow a \textit{truncated discrete distribution} $\text{TD} (\lambda,R)$ with a parameter $\lambda$ and a truncation range $R=[0,n)$. Alternatively, we write as truncation level $R=n$ if the left truncation is at zero in the non-negative case.
\end{definition}
%\begin{definition}\label{def_truncatable_dist_both}
%    For a discrete random variable $X \sim \text{D}(\lambda)$, define (1) $Z_{m,n} = X$ if $m < X < n$; (2) $Z_{m,n} = n-1$ if $X \geq n$; and (3) $Z_{m,n} = m+1$ if $X \leq m$.
%    The random variable $Z_{m,n}$ is said to follow a \textit{truncated discrete distribution} $\text{TD} (\lambda,R)$ with a parameter $\lambda$ and a truncation range $R=(m,n)$.
%\end{definition}
%\begin{definition}\label{def_truncatable_dist}
%    A \textit{truncated discrete random variable} $Z_{n}$ of a non-negative discrete random variable $X \sim \text{D}(\lambda)$ is a discrete random variable such that $Z_n = X$ if $X \leq n-1$, and $Z_n = n-1$ if $X > n$.
%    The random variable $Z_{n}$ is said to follow a \textit{truncated discrete distribution} $\text{TD} (\lambda,R)$ with a parameter $\lambda$ and a truncation range $R=[0,n)$. Alternatively, we write as truncation level $R=n$ if the left truncation is at zero in the non-negative case.
%\end{definition}
%\begin{definition}\label{def_truncatable_dist_both}
%    A \textit{truncated discrete random variable} $Z_{m,n}$ of a discrete random variable $X \sim \text{D}(\lambda)$ is a discrete random variable such that (1) $Z_{m,n} = X$ if $m < X < n$; (2) $Z_{m,n} = n-1$ if $X \geq n$; and (3) $Z_{m,n} = m+1$ if $X \leq m$.
%    The random variable $Z_{m,n}$ is said to follow a \textit{truncated discrete distribution} $\text{TD} (\lambda,R)$ with a parameter $\lambda$ and a truncation range $R=(m,n)$.
%\end{definition}

Definition \ref{def_truncatable_dist} not only finitizes the support but also provides the modification of the PMF: $\pi_k = P(Z_n=c_k) = P(X=c_k)$ for $k = 1,\cdots,n-1$, and $\pi_n = P(Z_n=c_n) = 1 - \sum_{k=1}^{n-1} \pi_k$.
%Algorithm \ref{algorithm_pmf} provides the computation on the modified PMF in Definition \ref{def_truncatable_dist}.
%Suppose we have a discrete random variable $X \sim \text{D}(\lambda)$ and its truncated random variable $Z \sim \text{TD} (\lambda,R)$, where $R$ denotes the truncation range that needs to be pre-defined.
%Then, we may now assume $Z$ has a support $C = \{ c_1, \cdots, c_{n} \}$ of $n$ possible outcomes and its corresponding constant outcome vector $c = (c_1, \cdots, c_{n})$.
For the truncated random variable $Z_n \sim \text{TD} (\lambda,R)$ having an outcome vector $c = ( c_1, \cdots, c_{n} )$, \textit{the ordering of $c_k$ is not significant, as long as the index of the outcome vector $c$ and the PMF vector $\pi$ are aligned.}
%In the case of Definition \ref{def_truncatable_dist_both}, we slightly abuse the definition when computing the PMF values.
Proposition \ref{thm_d_td} provides a theoretical basis that $Z_n$ approximates $X$ as the truncation range widens enough.
Hence, the truncation enables $\textsc{GenGS}$ to be extendedly applied to the discrete distributions with infinite support such as the Poisson, geometric, negative binomial, etc.
Similarly, Definition \ref{def_truncatable_dist} and Proposition \ref{thm_d_td} can be generalized to general discrete distribution as Proposition \ref{thm_d_td2} which we omit the proof.
\begin{proposition}\label{thm_d_td}
    %With Definition \ref{def_truncatable_dist}, 
    $Z_n$ \textit{converges} to $X$ \textit{almost surely} as $n \rightarrow \infty$.
\end{proposition}
\begin{proof}
    Note that $\{ \omega \in \Omega | Z_n (\omega) = X(\omega) \} = \{ \omega \in \Omega | X(\omega) < n \}$.
    Then, we have the following:
    \begin{align*}
    P( \lim_{n \rightarrow \infty} Z_n = X)
    & = P(\{ \omega \in \Omega | \lim_{n \rightarrow \infty} Z_n (\omega) = X(\omega) \}) \\
    & = P(\{ \omega \in \Omega | \lim_{n \rightarrow \infty} Z_n(\omega) = \lim_{n \rightarrow \infty} X (\omega) < \lim_{n \rightarrow \infty} n \} \}) \\
    & = P(\{ \omega \in \Omega | X(\omega) < \infty \}) = P(X < \infty) = 1
    \end{align*}
    since $X$ is a non-negative discrete random variable.%, and the cumulative distribution function $F_X (x)$ is non-decreasing with $\lim_{x \rightarrow \infty} F_X (x) = 1$.
\end{proof}
\begin{proposition}\label{thm_d_td2}
    For a discrete random variable $X \sim \text{D}(\lambda)$, define (\romannum{1}) $Z_{m,n} = X$ if $m < X < n$; (\romannum{2}) $Z_{m,n} = n-1$ if $X \geq n$; and (\romannum{3}) $Z_{m,n} = m+1$ if $X \leq m$.
    %The random variable $Z_{m,n}$ is said to follow a \textit{truncated discrete distribution} $\text{TD} (\lambda,R)$ with a parameter $\lambda$ and a truncation range $R=(m,n)$.
    Then, $Z_{m,n}$ \textit{converges} to $X$ \textit{almost surely} as $m \rightarrow -\infty$ and $n \rightarrow \infty$.
\end{proposition}

\section{Hyper-Parameters \& Variations of \textsc{GenGS}}\label{hpa}

Figure \ref{fig_graph_concept} illustrates the full steps of the \textsc{GenGS} trick, and Algorithm \ref{algorithm_pmf} shows the alternative sampling process of \textsc{GenGS}.
Figure \ref{fig_approximation} describes how the \textsc{GenGS} trick approximates the original distribution by adjusting the temperature and the truncation range.
The following are the hyper-parameter and the variations of the proposed \textsc{GenGS}.

\begin{algorithm}[t]
    \centering
    \caption{Alternative sampling of \textsc{GenGS} with explicit PMF.}
    \label{algorithm_pmf}
    \small
    \begin{algorithmic}[1]
    \STATE \textbf{Input:} Distribution $\text{D}(\lambda)$ with PMF $p_{\lambda} (\cdot)$, finite possible outcome vector $c$ within truncation range $R$, temperature $\tau$. 
%    \STATE \textbf{for} {$k=1$ {\bfseries to} $n-1$} \textbf{do}
    \STATE $\pi_k = p_{\lambda} (c_k)$ \textbf{for} {$k=1$ {\bfseries to} $n-1$}
    %\ENDFOR 
    \STATE $\pi_{n} = 1 - \sum_{k=1}^{n-1} \pi_k$
    %\STATE \textbf{return} $\pi=(\pi_1,\cdots,\pi_n)$
    \STATE Draw Gumbel samples $g_1, \cdots, g_n$.
    \STATE Compute ralaxed one-hot sample $w$ with $\textsc{GS} (\pi,\tau)$.
    \STATE Compute $z = \mathcal{T} (w) = \sum_{k=1}^{n} w \odot c$.
    \STATE \textbf{return } $z$
    \end{algorithmic}
\end{algorithm}

\paragraph{Temperature}
Since \textsc{GenGS} utilizes the GS, \textsc{GenGS} inherits the temperature as a hyper-parameter.
The decrement of temperature $\tau$ from results in the closer distribution to the original distribution.
However, the initially small $\tau$ leads to high bias and variance of gradients, which becomes problematic at the learning stage on $\pi$.
Hence, the annealing of $\tau$ from a large value to a small value is necessary to provide a learning chance of $\pi$.

\paragraph{Truncation Range}
The truncated distribution becomes closer to the original distribution as we widen the truncation range $R$, and the choice of truncation range is crucial in terms of covering many probable samples or the modes of the distribution. 
Hence, we can set the truncation range to cover most of the support through a certain threshold.
%, for example, except all but less than $\texttt{1e-5}\%$ of probability with respect to the prior distribution.
Also, for the popular discrete distributions such as Poisson, geometric, negative binomial, etc. which have a single mode in the PMF, we can use the mode as a criterion.
%However, requiring a pre-defined truncation range $R$ can be problematic since we cannot fully utilize the original distribution.
Alternatively, we can set non-parameteric truncation range by thresholding, as in Algorithm \ref{algorithm_pmf_nonparam}.
%However, it still suffers from over-dispersed distributions, since such distributions will require a large amount of time to compute PMF values.
%Hence, the choice of the thresholds depends on the distribution and is crucial in the implementation.

\begin{algorithm}[t]
    \centering
    \caption{\textsc{GenGS NP}: Computing PMF values without pre-defined truncation range.}
    \label{algorithm_pmf_nonparam}
    \small
    \begin{algorithmic}[1]
    \STATE \textbf{Input:} Distribution $\text{D}(\lambda)$ with and PMF $p_{\lambda} (\cdot)$, temperature $\tau$, threshold $\eta \lesssim 1$. %$\delta \gtrsim 0$ and 
    \STATE \texttt{k, prob$\_$sum} $= 0, 0$
    \STATE \texttt{pi, c = list(), list()}
    \STATE \textbf{while} {\texttt{prob$\_$sum} $< \eta$} \textbf{do}
        \STATE \hspace{1.em} \texttt{pi.append(}$p_{\lambda} (\texttt{k})$\texttt{), c.append(k)}
        \STATE \hspace{1.em} \texttt{prob$\_$sum, k +=}~ $p_{\lambda} (\texttt{k}), 1$
%        \STATE \hspace{1.em} \texttt{k += 1}
    %\ENDWHILE
    \STATE \texttt{pi.append(1-sum(pi)), c.append(k)}
    \STATE \textbf{return } \texttt{pi, c}
    \end{algorithmic}
\end{algorithm}

%\paragraph{Nonparameteric Truncation Range}
%We provide an alternative choice of pre-defined truncation range in Algorithm \ref{algorithm_pmf_nonparam}, and denote as \textsc{GenGS NP}.
%Although the alternative requires thresholds to stop computing PMF value at a certain level, Algorithm \ref{algorithm_pmf_nonparam} allows searching further meaningful range compared to the original \textsc{GenGS}.
%However, it still suffers from over-dispersed distributions, since such distributions will require a large amount of time to compute PMF values.
%Hence, the choice of the thresholds depends on the distribution and is crucial in the implementation.

\begin{algorithm}[t]
    \centering
    \caption{\textsc{GenGS Imp}: \textsc{GenGS} with implicit PMF.}
    \label{algorithm_imp}
    \small
    \begin{algorithmic}[1]
    \STATE \textbf{Input:} Distribution $\text{D}(\hat{\lambda})$ with inferred $\hat{\lambda}$, finite possible outcome vector $c$ within truncation range $R$, temperature $\tau$. 
    \STATE $\hat{\pi} = \texttt{softmax} (\hat{\lambda})$
    \STATE Compute $z = \textsc{GenGS} (\hat{\pi},\tau)$.
    \STATE \textbf{return } $z$
    \STATE $\hat{\lambda}$ is further optimized by Proposition \ref{theorem_kld}.
    \end{algorithmic}
\end{algorithm}

\paragraph{Implicit Inference}
The stochastic gradient estimators or the alternative sampling with the reparameterization tricks are widely utilized for the sampling from some distribution after infering the distribution parameter.
In such cases, there are distribution regularizer in the objective function, for example in VAEs, the KL divergence between the prior distribution and the approximate posterior distribution.
To do so, we first infer the distribution parameter $\lambda$, and then the proposed \textsc{GenGS} can be utilized by computing the PMF values $\pi$ explicitly.
However, we can instead directly infer the PMF values $\pi$ with \texttt{softmax} function, not through the distribution parameter $\lambda$.
%The proposed \textsc{GenGS} in Algorithm \ref{algorithm_explicit} enables inference on the distribution parameter $\lambda$ by computing the modified PMF values $\pi$ from the truncation.
%Meanwhile, we can directly infer the PMF values $\pi$ with \texttt{softmax} function, not through the distribution parameter $\lambda$, by loosening the assumption on the approximate posterior PMF shape.
We name the two cases as explicit inference and implicit inference (\textsc{GenGS Imp}), respectively, and \textsc{GenGS Imp} becomes possible by truncating distribution due to the finiteness of the dimension in the \texttt{softmax} function with the neural networks.
%However, this approach needs a regularizer, such as the KL divergence in the objective function of VAEs, which ensures the distribution shape to be similar to a prior distribution. 
Although the implicit inference cannot directly infer the distribution parameter $\lambda$, we found that loosening the posterior shape leads to a significant performance gain in our VAE examples.
Algorithm \ref{algorithm_imp} presents \textsc{GenGS Imp} process.

\paragraph{Discretization}
Straight-Through (ST) method \citep{Bengio13,Jang17} can be applied to \textsc{GenGS} for drawing discrete samples.
%We can apply the Straight-Through method \citep{Bengio13,Jang17} to \textsc{GenGS}, namely \textsc{GenGS ST}.
This variation samples discrete values in the feed-forward step, but utilizes the estimated gradient from the continuously relaxed and reparameterized $\textsc{GenGS}$ in the back-propagation step.
%\textsc{GenGS} outputs a continuously relaxed and reparameterized sample value.
%Utilizing the Straight-Through (ST) GS estimator \citep{Bengio13,Jang17}, we can obtain the discrete sample as well.
%The concept of the discretization is that using the sample from the original distribution $\text{D} (\lambda)$ at the feed-forward step, but utilizing the estimated gradient from the continuously relaxed and reparameterized distribution $\textsc{GenGS} (\pi,\tau)$ at the back-propagation step.
However, as in the original GS with the ST technique, \textsc{GenGS} with ST results in significant performance degradation in our synthetic examples.
%We alternatively recommend utilizing continuously reparameterized samples during the optimization, and evaluating with discrete samples, which we denote as \textsc{GenGS Disc}.

%\paragraph{Coverage}
%Theoretically, if a discrete distribution has an explicit PMF, then we can compute each PMF value, hence, the proposed \textsc{GenGS} can reparameterize the discrete distribution.

\begin{figure*}[t]
    \vspace{-.5em}
    \centering
    \begin{subfigure}
    \centering
    \includegraphics[width=.325\linewidth]{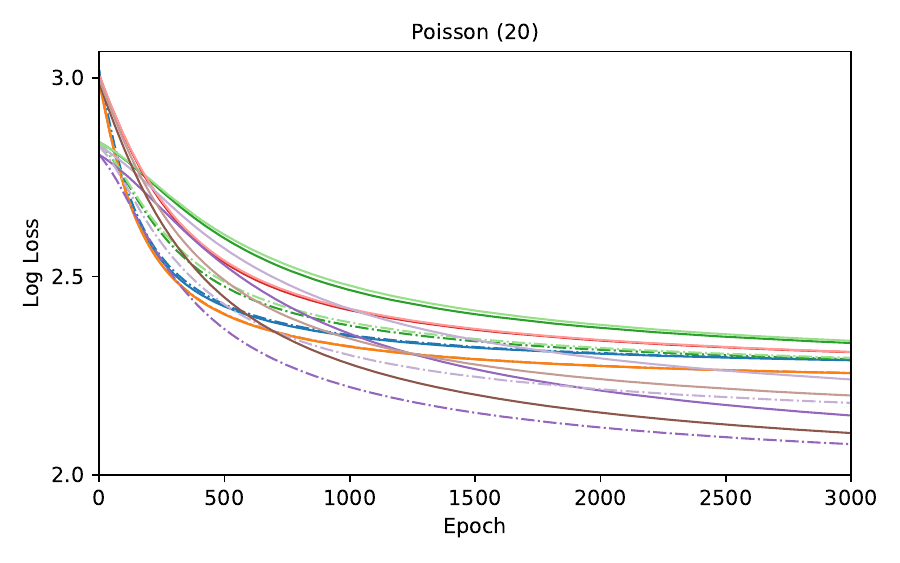}
    \end{subfigure}\vspace{-.45em}
    \begin{subfigure}
    \centering
    \includegraphics[width=.325\linewidth]{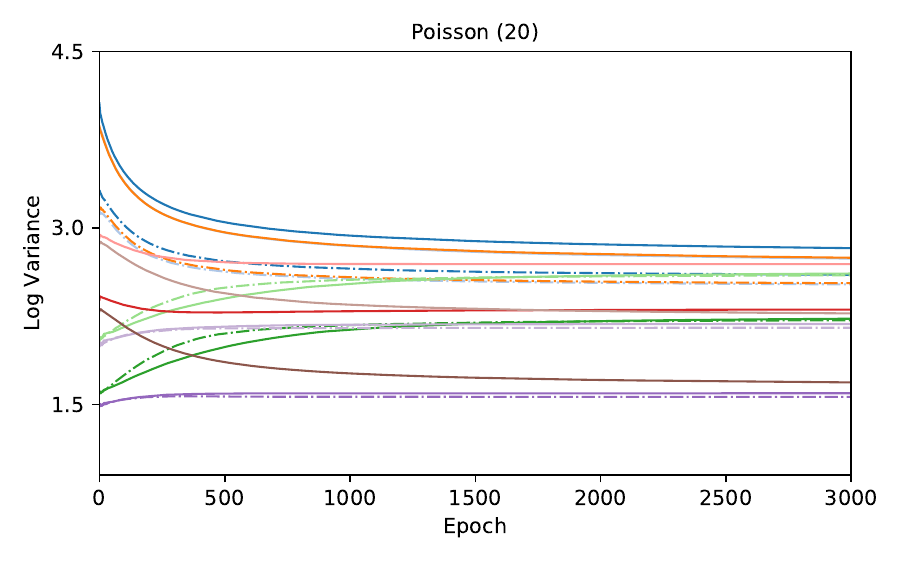}
    \end{subfigure}\vspace{-.45em}
    \begin{subfigure}
    \centering
    \includegraphics[width=.325\linewidth]{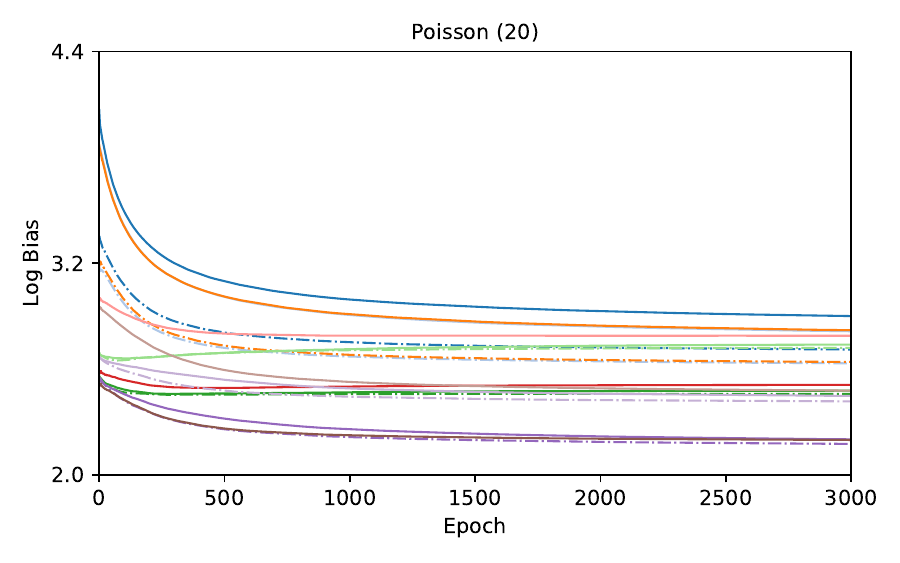}
    \end{subfigure}\vspace{-.45em}
    \begin{subfigure}
    \centering
    \includegraphics[width=.325\linewidth]{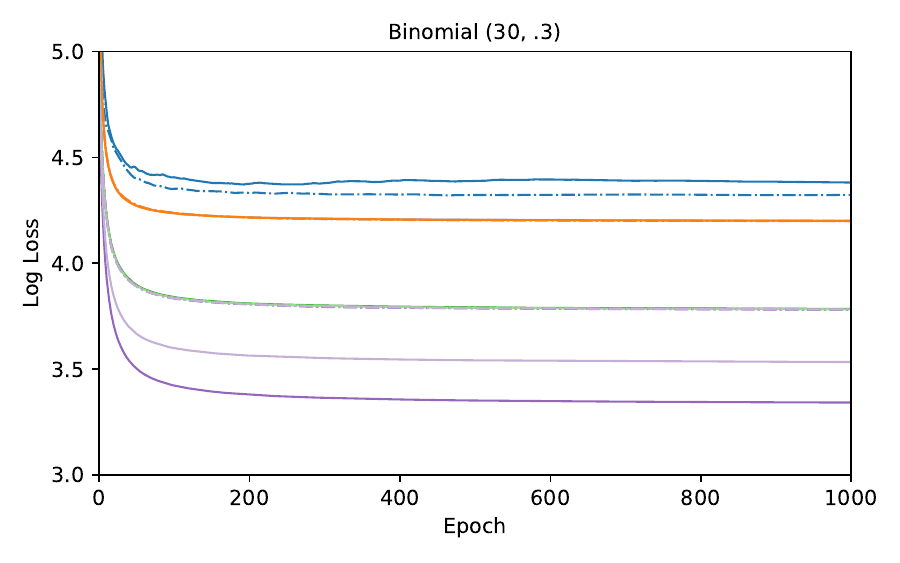}
    \end{subfigure}\vspace{-.225em}
    \begin{subfigure}
    \centering
    \includegraphics[width=.325\linewidth]{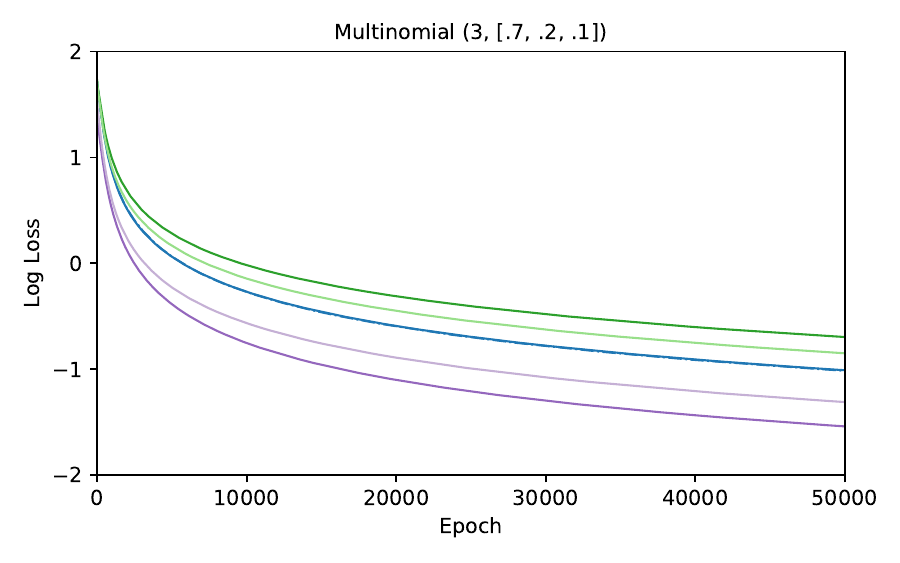}
    \end{subfigure}\vspace{-.225em}
    \begin{subfigure}
    \centering
    \includegraphics[width=.325\linewidth]{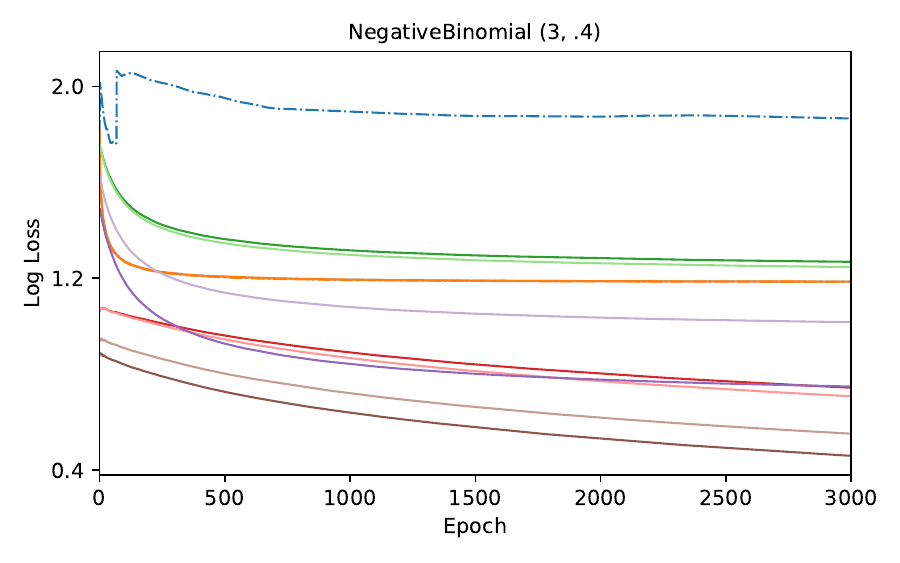}
    \end{subfigure}\vspace{-.225em}
    \begin{subfigure}
    \centering
    \includegraphics[width=.325\linewidth]{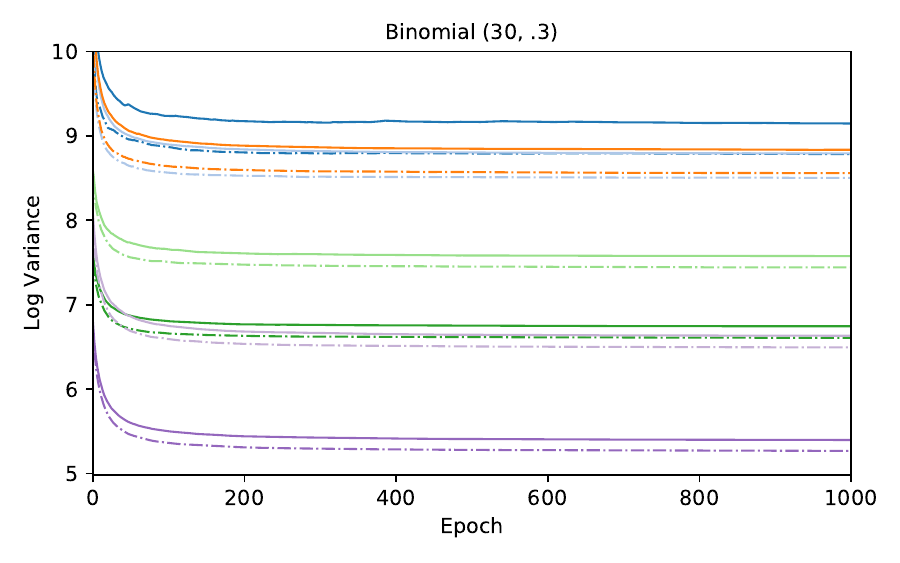}
    \end{subfigure}\vspace{-.3em}
    \begin{subfigure}
    \centering
    \includegraphics[width=.325\linewidth]{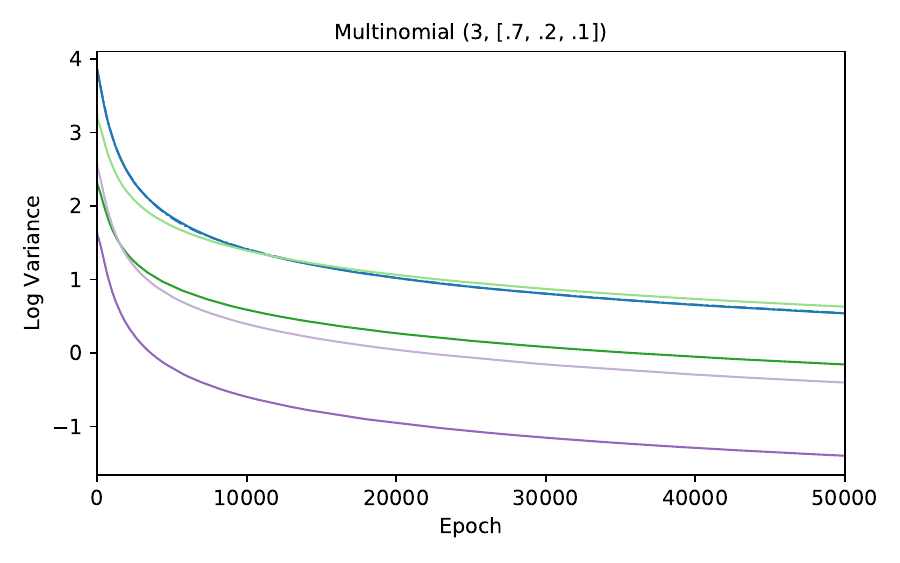}
    \end{subfigure}\vspace{-.3em}
    \begin{subfigure}
    \centering
    \includegraphics[width=.325\linewidth]{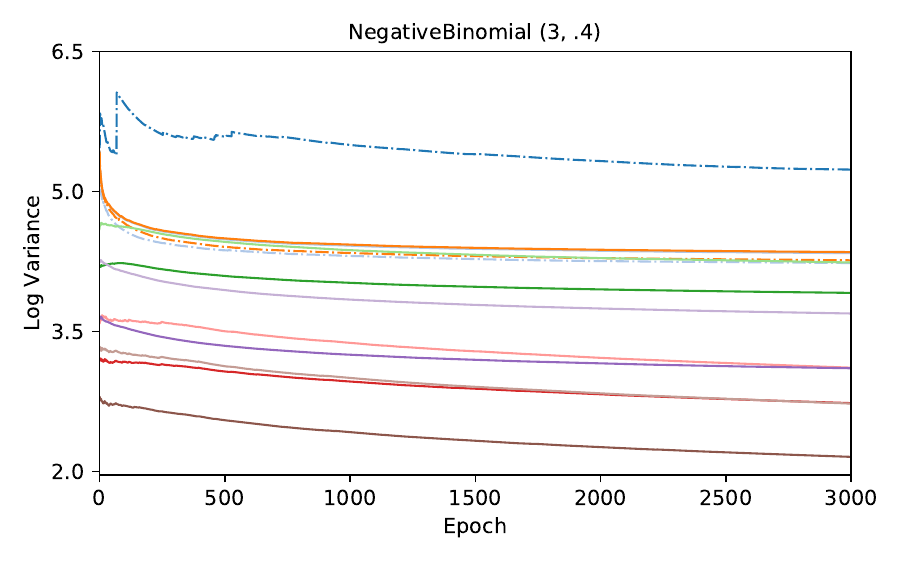}
    \end{subfigure}\vspace{-.3em}
    \begin{subfigure}
    \centering
    \includegraphics[width=.99\linewidth]{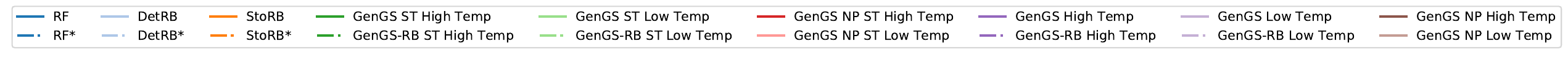}
    \end{subfigure}
    \vspace{-1.em}
    \caption{Synthetic example performance curves in \texttt{log} scale: (Top Row) losses, variances, and biases of gradients for $\text{Poisson}$; (Middle Row) losses for $\text{Binomial}$, $\text{Multinomial}$, and $\text{NegativeBinomial}$; (Bottom Row) variances of gradients for $\text{Binomial}$, $\text{Multinomial}$, and $\text{NegativeBinomial}$. We utilize the cumulative average for smoothing the curves following from \citet{Kool20}.}%, and the curves with confidence intervals and the curves without smoothing are in Appendix I.}
    \label{fig_toy}
    \vspace{-.6em}
\end{figure*}

\section{Related Work}\label{related_work}
\textsc{RF} denotes the basic REINFORCE \citep{Williams92}.
\textsc{NVIL} \citep{Mnih14} utilizes a neural network to introduce the optimal control variate.
\textsc{MuProp} \citep{Gu16} utilizes the first-order Taylor expansion on the objective function as a control variate.
\textsc{VIMCO} \citep{Mnih16} is designed as multi-sample gradient estimator.
\textsc{REBAR} \citep{Tucker17} and \textsc{RELAX} \citep{Grathwohl17} utilize reparameterization trick for constructing the control variate.
\textsc{DetRB} \citep{Liu19} uses the weighted value of the fixed gradients from $m$-selected categories and the estimated gradients from the remaining categories with respect to their odds to reduce the variance.
The idea of \textsc{StoRB} \citep{Kool20} is essentially same as that of \textsc{DetRB}, but \textsc{StoRB} randomly chooses the categories at each step. %, instead of using fixed categories.
\citet{Kool20} also suggested \textsc{UnOrd}, utilizing samples without replacements.
Hence, \textsc{DetRB}, \textsc{StoRB}, and \textsc{UnOrd} can be considered as multi-sample gradient estimators.
The $^*$ symbol denotes a built-in control variate introduced in the work of \citet{Kool20}.
\textsc{IRG} \citep{Potapczynski20} is an alternative of the GS and its variations can be applied to approximate the PMF including the Poisson, etc.
However, \textsc{IRG} does not infer the distribution parameter as \textsc{GenGS Imp}, and the paper mainly focuses on replacing the GS rather than reparameterizing the general class of the discrete random variables.

\begin{table*}[t]
\centering
\vspace{-.5em}
\caption{Training negative ELBO on MNIST and OMNIGLOT datasets. The lower is better for the negative ELBO, and the best and the second best results are marked in \textbf{bold} and \underline{underline}, repecitvely. The symbol ``---'' indicates no convergence.}
\label{table_vae}
\vspace{-.2em}
\resizebox{.99\textwidth}{!}{
    \begin{tabular}{lccccccccccccccc}
    \midrule
    ~ & \multicolumn{8}{c}{Single-Sample Gradient Estimators} & \multicolumn{6}{c}{Multi-Sample (10) Gradient Estimators} \\
    \cmidrule(r){2-9}
    \cmidrule(r){10-15}
    MNIST & \textsc{RF$^*$} & \textsc{NVIL} & \textsc{MuProp} & \textsc{REBAR} & \textsc{RELAX} & \textsc{GenGS} & \textsc{GenGS Imp} & \textsc{GenGS NP} & \textsc{VIMCO} & \textsc{StoRB$^*$} & \textsc{UnOrd} & \textsc{GenGS} & \textsc{GenGS Imp} & \textsc{GenGS NP} \\
    \midrule
    Pois(2) & $122.81_{\pm{2.41}}$ & $129.34_{\pm{4.72}}$ & $125.43_{\pm{2.27}}$ & $123.44_{\pm{2.54}}$ & $122.71_{\pm{1.92}}$ & $\underline{103.18}_{\pm{0.92}}$ & $\textbf{96.04}_{\pm{1.44}}$ & $104.98_{\pm{1.19}}$ & $122.13_{\pm{3.02}}$ & $122.71_{\pm{3.83}}$ & $130.95_{\pm{4.66}}$ & $\underline{99.28}_{\pm{1.36}}$ & $\textbf{94.78}_{\pm{1.17}}$ & $102.59_{\pm{1.32}}$ \\
    Pois(3) & $123.12_{\pm{2.21}}$ & $130.24_{\pm{3.32}}$ & $125.92_{\pm{1.81}}$ & $120.62_{\pm{2.31}}$ & $119.84_{\pm{2.18}}$ & $105.15_{\pm{1.71}}$ & $\textbf{96.01}_{\pm{1.27}}$ & $\underline{104.25}_{\pm{1.56}}$ & $120.93_{\pm{2.48}}$ & $121.92_{\pm{3.12}}$ & $119.97_{\pm{5.38}}$ & $\underline{100.17}_{\pm{1.73}}$ & $\textbf{94.57}_{\pm{1.35}}$ & $105.51_{\pm{1.80}}$ \\
    Geom(.25) & $127.90_{\pm{1.97}}$ & $135.90_{\pm{2.38}}$ & $137.90_{\pm{2.14}}$ & $135.12_{\pm{2.74}}$ & $136.80_{\pm{3.06}}$ & $\underline{98.43}_{\pm{0.81}}$ & $\textbf{92.52}_{\pm{1.62}}$ & $101.77_{\pm{1.47}}$ & $126.46_{\pm{2.22}}$ & $130.96_{\pm{3.36}}$ & --- & $\underline{92.59}_{\pm{1.07}}$ & $\textbf{92.06}_{\pm{0.78}}$ & $100.34_{\pm{1.77}}$ \\
    Geom(.5) & $129.20_{\pm{2.03}}$ & $138.47_{\pm{2.30}}$ & $136.40_{\pm{1.78}}$ & $138.37_{\pm{2.98}}$ & $139.41_{\pm{3.59}}$ & $100.92_{\pm{1.24}}$ & $\textbf{93.81}_{\pm{1.60}}$ & $\underline{97.29}_{\pm{1.89}}$ & $128.71_{\pm{2.81}}$ & $134.10_{\pm{4.65}}$ & --- & $98.27_{\pm{1.22}}$ & $\textbf{91.46}_{\pm{0.95}}$ & $\underline{96.44}_{\pm{1.38}}$ \\
    NB(3,.5) & $116.67_{\pm{5.97}}$ & $119.28_{\pm{7.80}}$ & $131.96_{\pm{6.49}}$ & --- & --- & $\underline{98.58}_{\pm{1.27}}$ & $\textbf{94.52}_{\pm{1.52}}$ & $107.14_{\pm{2.14}}$ & $112.11_{\pm{3.19}}$ & $110.52_{\pm{2.21}}$ & --- & $\underline{96.83}_{\pm{1.63}}$ & $\textbf{93.66}_{\pm{1.03}}$ & $99.85_{\pm{1.91}}$ \\
    NB(5,.3) & $130.03_{\pm{3.99}}$ & $133.44_{\pm{4.27}}$ & $144.05_{\pm{8.15}}$ & --- & --- & $\underline{100.88}_{\pm{2.35}}$ & $\textbf{95.37}_{\pm{1.43}}$ & $110.41_{\pm{2.81}}$ & $121.29_{\pm{2.93}}$ & $127.32_{\pm{4.79}}$ & --- & $\underline{97.29}_{\pm{1.87}}$ & $\textbf{94.29}_{\pm{1.28}}$ & $105.52_{\pm{2.53}}$ \\
    \midrule
    OMNIGLOT & \textsc{RF$^*$} & \textsc{NVIL} & \textsc{MuProp} & \textsc{REBAR} & \textsc{RELAX} & \textsc{GenGS} & \textsc{GenGS Imp} & \textsc{GenGS NP} & \textsc{VIMCO} & \textsc{StoRB$^*$} & \textsc{UnOrd} & \textsc{GenGS} & \textsc{GenGS Imp} & \textsc{GenGS NP} \\
    \midrule
    Pois(2) & $139.47_{\pm{3.29}}$ & $148.01_{\pm{4.19}}$ & $142.95_{\pm{1.32}}$ & $138.12_{\pm{3.26}}$ & $137.56_{\pm{2.94}}$ & $127.89_{\pm{1.44}}$ & $\textbf{118.17}_{\pm{2.22}}$ & $\underline{123.40}_{\pm{1.71}}$ & $138.21_{\pm{3.29}}$ & $137.87_{\pm{4.31}}$ & $146.60_{\pm{4.97}}$ & $\underline{120.42}_{\pm{2.21}}$ & $\textbf{116.92}_{\pm{1.52}}$ & $121.81_{\pm{2.47}}$ \\
    Pois(3) & $140.54_{\pm{2.36}}$ & $148.13_{\pm{3.98}}$ & $143.85_{\pm{1.54}}$ & $137.92_{\pm{3.07}}$ & $137.42_{\pm{2.96}}$ & $131.53_{\pm{1.76}}$ & $\textbf{119.15}_{\pm{1.92}}$ & $\underline{126.03}_{\pm{2.16}}$ & $138.84_{\pm{2.69}}$ & $139.89_{\pm{3.98}}$ & $147.18_{\pm{5.23}}$ & $\underline{121.93}_{\pm{2.49}}$ & $\textbf{117.71}_{\pm{1.28}}$ & $124.71_{\pm{2.16}}$ \\
    Geom(.25) & $142.68_{\pm{2.96}}$ & $153.69_{\pm{2.52}}$ & $152.17_{\pm{1.77}}$ & $146.78_{\pm{3.62}}$ & $148.91_{\pm{4.03}}$ & $\underline{115.23}_{\pm{2.00}}$ & $\textbf{107.79}_{\pm{2.84}}$ & $116.75_{\pm{1.90}}$ & $142.42_{\pm{3.61}}$ & $142.54_{\pm{3.21}}$ & --- & $\underline{113.49}_{\pm{1.95}}$ & $\textbf{105.38}_{\pm{1.13}}$ & $115.69_{\pm{2.03}}$ \\
    Geom(.5) & $142.70_{\pm{1.77}}$ & $153.20_{\pm{1.49}}$ & $149.76_{\pm{2.19}}$ & $149.63_{\pm{3.49}}$ & $151.97_{\pm{3.90}}$ & $115.14_{\pm{2.43}}$ & $\textbf{108.48}_{\pm{2.78}}$ & $\underline{112.76}_{\pm{2.57}}$ & $141.52_{\pm{3.38}}$ & $140.68_{\pm{4.61}}$ & --- & $\underline{108.91}_{\pm{1.89}}$ & $\textbf{105.97}_{\pm{1.30}}$ & $112.41_{\pm{1.85}}$ \\
    NB(3,.5) & $141.44_{\pm{2.20}}$ & $144.44_{\pm{2.78}}$ & $147.78_{\pm{4.49}}$ & --- & --- & $\underline{118.57}_{\pm{2.71}}$ & $\textbf{117.02}_{\pm{2.18}}$ & $123.96_{\pm{2.38}}$ & $141.22_{\pm{3.42}}$ & $128.66_{\pm{3.88}}$ & --- & $\underline{116.55}_{\pm{2.23}}$ & $\textbf{114.20}_{\pm{1.61}}$ & $120.92_{\pm{2.34}}$ \\
    NB(5,.3) & $145.16_{\pm{3.83}}$ & $159.40_{\pm{5.13}}$ & $152.81_{\pm{3.34}}$ & --- & --- & $\underline{119.57}_{\pm{2.02}}$ & $\textbf{117.54}_{\pm{2.76}}$ & $127.46_{\pm{3.51}}$ & $148.89_{\pm{4.51}}$ & --- & --- & $\underline{118.37}_{\pm{2.51}}$ & $\textbf{114.77}_{\pm{1.47}}$ & $124.98_{\pm{2.95}}$ \\
    \midrule
\vspace{-1.2em}
\end{tabular}
}
\end{table*}

\section{Experiment}\label{experiment}
%We use (1) Intel Core i7-6700K CPU, 32GB RAM, and Titan X \& P40 GPUs; and (2) TensorFlow 1.15.0, TensorFlow Probability 0.8.0, and PyTorch 1.0.1 \& 1.2.0.
%Also, we run the experiments over $10$ times for each experiment.
%\textsc{GenGS} indicates the proposed method where additional terms \textsc{ST}, \textsc{Disc}, \textsc{Imp}, and \textsc{NP} indicates the variations, described in Section \ref{hpa}.

\subsection{Synthetic Example}\label{experiment_toy}
\paragraph{Experimental Setting}
This experiment expands the toy experiments from \citet{Tucker17,Grathwohl17} to diverse discrete distributions.
We optimize the objective function $\mathds{E}_{z \sim p(z|\lambda)} \big[ \sum_{i=1}^k (z_i - t_i)^2 \big]$ with respect to parameter $\lambda$ for fixed constants $t_1, \cdots, t_k$.
%First, we fix constant $t_1, \cdots, t_k$, and then optimize the objective function $\mathds{E}_{z \sim p(z|\lambda)} \big[ \sum_{i=1}^k (z_i - t_i)^2 \big]$ with respect to parameter $\lambda$.
Here, we set distribution $p(z|\lambda)$ as $\text{Poisson} (\lambda)$, $\text{Binomial} (20,\lambda)$, $\text{Multinomial} (3,\lambda)$, and $\text{NegativeBinomial} (3,\lambda)$.
%Here, we set distribution $p(z|\lambda)$ as $\text{Poisson} (\lambda=20)$, $\text{Binomial} (20,\lambda=.3)$, $\text{Multinomial} (3,\lambda=[.7,.2,.1])$, and $\text{NegativeBinomial} (3,\lambda=.4)$.
For the Poisson and the negative binomial, which have infinite supports, we also applied Algorithm \ref{algorithm_pmf_nonparam} to search the full range of support without pre-defined truncation range.
Since the \textsc{GenGS}es are basically a single-sample gradient estimator, we utilize a single sample of $z$ for the fundamental comparison among the gradient estimators in the toy example.

\paragraph{Experimental Result}
Figure \ref{fig_toy} compares the log-loss and the log-variance of estimated gradients from various estimators.
The log-loss needs to be minimized with the estimated gradient value in the learning process by back-propagation. 
Additionally, the log-variance requires being minimized to maintain the consistency of the gradients, so the gradient descent can be efficient.
The \textsc{GenGS}es show the best log-loss and the best log-variance if the \textsc{GenGS}es keep the continuous relaxation of the modeled discrete random variable.
For the Poisson, the exact gradient can be computed in closed-form, and the \textsc{GenGS}es show the lowest bias among all gradient estimators.

\subsection{VAE: Synthetic Experiment on DGMs}\label{experiment_vae}
\paragraph{Experimental Setting}
To test the performance of the gradient estimators in DGMs, we adopt VAE which is one of the simplest DGMs.
We follow the VAE experiment scheme of \citet{Figurnov18} which utilizes various prior distributions.
In our discrete case, we utilize the Poisson (Pois), the geometric (Geom), and the negative binomial (NB) distributions, as the \textit{latent factor count}. 
Note that the purpose of the VAE experiments is \textit{not to compare the performance across various prior distributions} such as the categorical or the Gaussian, but \textit{to compare the performance across gradient estimators within the same prior distributions}.
The VAE is considered as a more challenging task than the synthetic example, since (1) this task requires computing the gradients of the encoder network parameters through the latent distribution parameter $\lambda$; and (2) each stochastic gradient of the latent dimension affects every encoder parameter since we are utilizing the fully-connected layers.
Hence, a single poorly estimated gradient of the latent distribution parameter $\lambda$ could harm the learning of encoder parameters, so the VAE experiment can dynamically show the performance of the gradient estimators.
Objective function of the VAE is the evidence lower bound (ELBO) $\mathcal{L} = \mathbb{E}_{q_{\phi (\bold{z}|\bold{x})}} [\log p_{\theta} (\bold{x}|\bold{z})] - \text{KL}(q_{\phi} (\bold{z}|\bold{x}) || p_{\theta} (\bold{z}))$. %, consists of the reconstruction part and the KL term.
In the \textsc{GenGS}es, by truncating the original distribution, the KL divergence between the approximate posterior and the prior distributions becomes the derivation with categorical distributions.
\begin{proposition}\label{theorem_kld}
    Assume two truncated distributions $X \sim \text{TD}(\lambda,n)$ and $Y \sim \text{TD} (\hat{\lambda},n)$ where $\pi_k = P(X=k)$, $\hat{\pi}_k = P(Y=k)$. Then, the KL divergence between $X$ and $Y$ can be represented in the KL divergence between the categorical distributions where $\text{KL} (Y || X) = \text{KL} (\text{Categorical}(\hat{\pi}) || \text{Categorical}(\pi))$.
\end{proposition}
\begin{proof}
    \begin{align*}
    \text{KL} (Y || X) & = \textstyle \sum_k P(Y=k) \log{ \bigg( \frac{P(Y=k)}{P(X=k)} \bigg)} = \textstyle \sum_k \hat{\pi}_k \log{ \Big( \frac{\hat{\pi}_k}{\pi_k} \Big)} \\
    & = \text{KL} (\text{Categorical}(\hat{\pi}) || \text{Categorical}(\pi))
    \end{align*}
\end{proof}
We separately compare the performance of the single-sample and multi-sample estimators with $10$ samples.
For the \textsc{GenGS}es, we anneal the temperature from $1.$ to $.1$, and set truncation levels $12$ for $\text{Pois} (2)$; $15$ for $\text{Pois} (3)$; $25$ for $\text{Geom} (.25)$; $15$ for $\text{Geom} (.5)$; $30$ for $\text{NB} (3,.5)$; and $30$ for $\text{NB} (5,.3)$.
%In the multi-sample case, we set truncation level as $50$ for all \textsc{GenGS} experiments.

\paragraph{Experimental Result}
To compare the estimators in the optimization perspective, Table \ref{table_vae} provides the training negative ELBO curves on MNIST and OMNIGLOT.
% of the VAE experiments.
\textsc{DetRB}, \textsc{DetRB$^*$}, and \textsc{StoRB} are not listed since they hardly lead to the optimal point.
The variants of \textsc{GenGS} showed the lowest negative ELBO in general for both the single and the multi cases.
%We found that as we anneal the temperature into value, the performance gap between \textsc{GenGS} and \textsc{GenGS Disc} reduces.
Also, loosening the PMF condition (i.e., the implicit inference) reached the better optimal points.
The empirical reason why the implicit version is better than the explicit version is that the posterior PMF shape in the implicit case is thinner.
Hence, the implicit distribution has a lower variance than the explicit one, and samples consistent values that lead to better trained neural network parameters, although it loses the original PMF shape.
%Figure \ref{fig_reconstruction} shows the reconstructed images by VAEs with various gradient estimators on MNIST and OMNIGLOT.
%\textsc{GenGS} draws the clearest images and better reconstructions, which aligns with the quantitative result in Table \ref{table_vae}.

\begin{figure}[t]
    \centering
    \includegraphics[width=.25\linewidth]{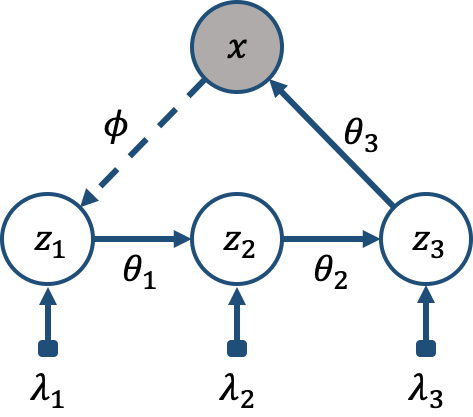}
    \hspace{2.em}
    \includegraphics[width=.4\linewidth]{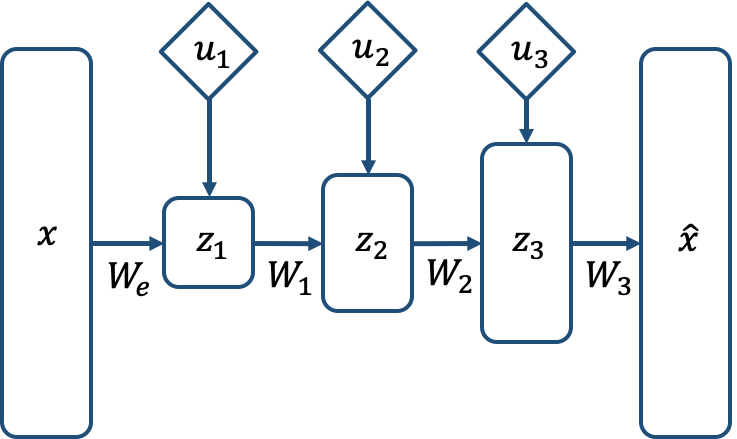}
    \vspace{-.4em}
    \captionof{figure}{(Left) A graphical notation of NVPDEF with the generative ($\theta$) and the inference ($\phi$) processes. (Right) A neural network view of NVPDEF: diamond nodes indicate the auxiliary random variable for the reparameterization trick.}
    \label{fig_nvpdef}
    \vspace{-.6em}
\end{figure}

\subsection{Application on Topic Model}\label{experiment_topic_model}
\paragraph{Experimental Setting}
This experiment shows practical usage of discrete distribution with \textsc{GenGS} in the deep generative topic modeling.
%Poisson distribution is one of the most important distributions for counting the number of outcomes among all discrete distributions.
The authors of \textit{Deep Exponential Families} (DEFs) \citep{Ranganath15} utilized the exponential family on the stacked latent layers.
We focus on the Poisson DEF, which assumes the Poisson latent layers to capture the count of latent super-topics and sub-topics.
We convert the Poisson DEF into a neural variational form, which resembles NVDM \citep{Miao16}, namely \textsc{NVPDEF}.
%the neural variational Poisson DEF (\textsc{NVPDEF}).

The generative and the inference process are
\begin{gather*}
    \text{Gen: } z_k \sim \text{Pois} (\lambda_{k-1}) \text{ for } k = 1, \cdots, K, \text{ and } x \sim \text{MLR} (\lambda_K) \\
    %\text{Gen: } z_1 \sim \text{Pois} (\lambda_0), \cdots, z_K \sim \text{Pois} (\lambda_{K-1}),\\
    %x \sim \text{MultinomialLogisticRegression} (\lambda_K) \\
    \text{Inf: } \hat{\lambda}_0 = \text{MLP} (x),~ \hat{\lambda}_1 = W_1 \hat{\lambda}_0, ~\cdots, ~\hat{\lambda}_K = W_{K-1} \hat{\lambda}_{K-1} ~,
\end{gather*}
where MLR stands for multinomial logistic regression adopted from \citet{Miao16}. 
%Here, we adopt multinomial logistic regression from NVDM \citep{Miao16}.
Each $z_k \sim \text{Poisson}(\lambda_{k-1})$ represents the count distribution of sub-topics from the super-topic, which is approximated by posterior $q(z_k|z_{k-1}) = \text{Pois}(\hat{\lambda}_k)$.
Each component of $W_k$, $w_{k,i,j}$ is positive, and $w_{k,i,j}$ captures the positive weight of relationship between super-topic $i$ of the $k^\text{th}$ layer and sub-topic $j$ of the $(k+1)^\text{th}$ layer.
Finally, \textsc{NVPDEF} optimizes
\begin{align}
    \mathcal{L} = \mathds{E}_{q(z_{1:K})} [\log p(x|z_{1:K})] \textstyle - \sum_{k=1}^K \text{KL} (q(z_k|z_{k-1})||p(z_k))
\end{align}
where $z_0 = x$ for simplifying the equation.
Figure \ref{fig_nvpdef} shows the graphical notation and the neural network structure.

For the single-stacked version of \textsc{NVPDEF}, we set $\lambda_1=.75$ with truncation level $15$ and fix the temperature as $\tau = .5$.
For the multi-stacked version of \textsc{NVPDEF}, we set $\lambda_1 = 1.1, \lambda_2 = 1.$ with truncation level $15$.
To have better chances of learning during the training period in the multi case, we utilize temperature annealing from $\tau = 3.$ to $\tau = .5$, and multi-sample on the latent layers for the stable optimization of consecutive sampling.
We compare the estimators by perplexity $\exp (- \frac{1}{D} \sum_d \frac{\log p(d)}{N_d})$ where $N_d$ is the number of words in document $d$, and $D$ is the total number of documents.

\begin{table}[t]
\vspace{-.5em}
\captionof{table}{\textsc{NVPDEF} test perplexity on 20Newsgroups and RCV1. $(m)$ in the model name indicates that $m$ samples are used for estimating the gradients. The lower is better for the perplexity, and the best results are marked in \textbf{bold}. The symbol ``---'' indicates no convergence.}%: (1) \textsc{DetRB} does not lead to the optimal point; and (2) \textsc{DetRB$^*$} and \textsc{StoRB} outperformed by \textsc{StoRB$^*$}.
\label{table_def}
\vspace{-.2em}
\centering
\resizebox{.87\linewidth}{!}{
    \begin{tabular}{lcclcc}
    \midrule
    \multirow{2.5}{*}{\shortstack[l]{Gradient \\ Estimator}} & \multicolumn{2}{c}{\textsc{1-Stacked} (Dim.)} & \multirow{2.5}{*}{\shortstack[l]{Gradient \\ Estimator}} & \multicolumn{2}{c}{\textsc{2-Stacked} (Dim.)} \\
    \cmidrule(r){2-3}
    \cmidrule(r){5-6}
     & 20News ($50$) & RCV1 ($200$) & & 20News ($50$) & RCV1 ($200$) \\
    \cmidrule(r){1-3}
    \cmidrule(r){4-6}
    \textsc{RF} & $1227_{\pm{33.4}}$ & $1182_{\pm{27.8}}$ & \textsc{RF}$(10)$ & --- & --- \\
    \textsc{RF$^*$} & $944_{\pm{17.7}}$ & $708_{\pm{15.6}}$ & \textsc{RF$^*$}$(10)$ & $1221_{\pm{19.6}}$ & $2190_{\pm{30.9}}$ \\
    \textsc{NVIL} & $1077_{\pm{22.5}}$ & $1191_{\pm{25.8}}$ & \textsc{NVIL}$(10)$ & --- & --- \\
    \textsc{MuProp} & $1045_{\pm{20.1}}$ & $935_{\pm{19.3}}$ & \textsc{MuProp}$(10)$ & --- & --- \\
    \textsc{VIMCO}$(10)$ & $958_{\pm{17.1}}$ & $741_{\pm{16.2}}$ & \textsc{VIMCO}$(20)$ & --- & --- \\
    \textsc{REBAR} & $934_{\pm{27.6}}$ & $716_{\pm{25.9}}$ & \textsc{REBAR}$(10)$ & $1136_{\pm{26.3}}$ & $2075_{\pm{43.6}}$ \\
    \textsc{RELAX} & $932_{\pm{31.1}}$ & $717_{\pm{29.4}}$ & \textsc{RELAX}$(10)$ & $1116_{\pm{23.1}}$ & --- \\
    \textsc{StoRB$^*$}$(10)$ & $919_{\pm{16.2}}$ & $701_{\pm{14.7}}$ & \textsc{StoRB$^*$}$(20)$ & $1094_{\pm{35.7}}$ & --- \\
    \textsc{UnOrd}$(10)$ & $1206_{\pm{24.3}}$ & $1188_{\pm{22.1}}$ & \textsc{UnOrd}$(20)$ & --- & --- \\
    \textsc{GenGS} & $\textbf{759}_{\pm{13.1}}$ & $\textbf{562}_{\pm{11.5}}$ & \textsc{GenGS}$(10)$ & $\textbf{783}_{\pm{17.6}}$ & $\textbf{576}_{\pm{18.8}}$ \\
    \midrule
    \vspace{-2.3em}
    \end{tabular}
    }
\end{table}

\paragraph{Experimental Result}
We enumerate the results of \textsc{NVPDEF} with various gradient estimators in Table \ref{table_def}, and we confirmed that \textsc{GenGS} gives the lowest perplexity with 20Newsgroups and RCV1, which is aligned with the VAE results. 
Especially, the multi-stacked RCV1 case extremely shows the performance of the gradient estimators that many of the gradient estimators fail to reach the optimal point and \textsc{GenGS} gives the lowest perplexity.

\section{Conclusion}\label{conclusion}
This paper suggests variants of \textsc{GenGS}, a generalized version of the Gumbel-Softmax estimator, with the theoretical background. 
%We utilize the truncation and the linear transformation to generalize the popularly used Gumbel-Softmax, and the proposed \textsc{GenGS} extends the range of reparameterizable discrete distributions.
% that can be reparameterized.
The synthetic analysis and the VAE experiment demonstrate the efficacy of \textsc{GenGS}, and the topic model application shows the usage of \textsc{GenGS}.
%Although it requires careful hyper-parameter selection on the truncation range and the temperature which can be regarded as limitations, 
The proposed \textsc{GenGS}es can be simply implemented, and the experimental result shows that the variants of \textsc{GenGS} lead to the better optimal point compared to existing stochastic gradient estimators. 
%Although the truncation range and the temperature need to be carefully tuned as hyper-parameters (if we use pre-defined truncation range), the experimental result shows that the variants of \textsc{GenGS} lead to the better optimal point compared to existing stochastic gradient estimators. 
With the generalization, we expect that \textsc{GenGS} can diversify the options of distributions in the deep generative model community.
%While DGMs can be used for generating vicious data, however, it is beyond the scope of this paper.

\section*{Declaration of Competing Interest}
The authors declare that there are no conflicts of interest.

\section*{Acknowledgments}
This work was supported by the National Research Foundation of Korea (NRF) grant funded by the Korea government (MSIT) (RS-2022-00166289).

\bibliographystyle{elsarticle-num-names}
\bibliography{references.bib}

\end{document}